\documentclass[letterpaper]{article} 
\usepackage{aaai24}  
\usepackage{times}  
\usepackage{helvet}  
\usepackage{courier}  
\usepackage[hyphens]{url}  
\usepackage{graphicx} 
\usepackage{natbib}  
\usepackage{caption} 
\usepackage{algorithm}
\usepackage{algorithmic}

\usepackage{newfloat}
\usepackage{listings}
\DeclareCaptionStyle{ruled}{labelfont=normalfont,labelsep=colon,strut=off} 
\floatstyle{ruled}
\newfloat{listing}{tb}{lst}{}
\floatname{listing}{Listing}
\title{From Cloud to Edge: Rethinking Generative AI for Low-Resource Design Challenges}
\author {
    Sai Krishna Revanth Vuruma\textsuperscript{\rm 1},
    Ashley Margetts\textsuperscript{\rm 2},
    Jianhai Su\textsuperscript{\rm 3},
    Faez Ahmed\textsuperscript{\rm 2},
    Biplav Srivastava\textsuperscript{\rm 3}
}
\affiliations {
    \textsuperscript{\rm 1}
    Department of Computer Science and Engineering, University of South Carolina, Columbia, South Carolina, USA\\
    
    \textsuperscript{\rm 2}
    Department of Mechanical Engineering, Massachusetts Institute of Technology, Cambridge, Massachusetts, USA\\

    \textsuperscript{\rm 3}
    AI Institute, University of South Carolina, Columbia, South Carolina, USA\\

    svuruma@email.sc.edu, amargett@mit.edu, suj@email.sc.edu, faez@mit.edu, biplav.s@sc.edu
}

\usepackage{bibentry}

\usepackage{comment}
\usepackage{booktabs}

\usepackage{color}

\begin{document}
\maketitle

\begin{abstract}
Generative Artificial Intelligence (AI) has shown tremendous prospects in all aspects of technology, including design. However, due to its heavy demand on resources, it is usually trained on large computing infrastructure and often made available as a cloud-based service. In this position paper, we consider the potential, challenges, and promising approaches for generative AI for design on the edge, i.e., in resource-constrained settings where memory, compute, energy (battery) and network connectivity may be limited. 

Adapting generative AI for such settings involves overcoming significant hurdles, primarily in how to streamline complex models to function efficiently in low-resource environments. This necessitates innovative approaches in model compression, efficient algorithmic design, and perhaps even leveraging edge computing. 
The objective is to harness the power of generative AI in creating bespoke solutions for design problems, such as medical interventions, farm equipment maintenance, and educational material design, tailored to the unique constraints and needs of remote areas. These efforts could democratize access to advanced technology and foster sustainable development, ensuring universal accessibility and environmental consideration of AI-driven design benefits.
\end{abstract}

\section{Introduction}
In the rapidly evolving landscape of engineering design, the integration of Artificial Intelligence (AI) has catalyzed a transformative shift. Generative AI has been shown to revolutionize design processes \cite{regenwetter2022deep} and optimize functional performance across diverse domains ranging from fashion \cite{Sbai_2018_ECCV_Workshops} to medicine \cite{walters2020assessing}. However, a crucial aspect often overlooked in this digital renaissance is the need for AI accessibility in remote or resource-constrained environments --- a realm where tiny, offline machine learning (ML) models may be needed.

The need for tiny AI models in remote areas exists for two main reasons. Firstly, these regions frequently face challenges such as limited internet connectivity and inadequate computational resources. In such scenarios, cloud-dependent large AI models are impractical. Tiny ML models tailored for offline use can overcome these barriers, bringing the power of advanced design tools to isolated communities. This democratization not only fuels local innovation but also ensures that cutting-edge AI-assisted design solutions are not the exclusive domain of well-resourced urban centers.

Unique challenges in applications such as medicine, agriculture, and education in remote areas could greatly benefit from compact, offline AI models. In medical intervention, AI-driven design tools could assist in creating medical devices, such as a ventilator, tailored for limited-resource settings, focusing on affordability and ease of use while ensuring high efficacy. 
For the repair of farm equipment such as irrigation pumps, AI can play a pivotal role in predicting equipment failures or guiding repairs, adapted to the specific machinery and agricultural practices of different regions. In the context of school supplies, AI models could help design educational materials like toys \cite{ccl-toys} that are not only engaging and culturally relevant but also accessible to students with limited resources.

Secondly, engineering design problems in remote areas often possess unique contextual nuances - from local material constraints to specific environmental considerations and product usage patterns. Offline ML models, trained on relevant local data, can better capture and respond to these nuances, offering more tailored and effective design solutions in spaces such as reliable sewage repair or low-cost greenhouse design. 

Moreover, these small AI models align with the principles of sustainable technology deployment. By reducing reliance on large data centers and continuous internet connectivity, they contribute to lower energy consumption and a smaller carbon footprint, which is particularly crucial in ecologically sensitive remote areas \cite{prakash2023tinyml}.

The challenge lies in distilling the complexity of generative models into lightweight versions without significant loss of functionality and performance. This requires innovative approaches in model compression, efficient algorithm design, and perhaps the use of edge computing architectures.
\section{Review of Current Models}
Generative AI techniques revolutionize the way machines can learn and replicate human behavior. They use Machine Learning (ML) techniques to create new data that is realistic and contextually relevant. Advances in Generative AI, combined with the introduction of the transformer architecture with BERT \cite{devlin2019bert} led to the development of Large Language Models (LLMs).

Prominent LLMs such as Meta's LLAMA \cite{touvron2023llama}, OpenAI's ChatGPT built on top of InstructGPT \cite{ouyang2022training}, have demonstrated remarkable capabilities in understanding and generating human-like text with applications ranging from sentiment analysis to chatbots. However, current models are primarily limited to a few modalities (such as text-based conversational agents). 

Visual input is key to understanding the nuances of generating and understanding designs. As a result, any potential Generative AI solution must be able to process images as input and output. One potential solution to this problem is Diffusion models such as OpenAI's DALL-E \cite{ramesh2021zeroshot}, Stability AI's Stable Diffusion \cite{rombach2022highresolution} or one of their more recent models SDXL \cite{podell2023sdxl}, which have been shown to generate high quality images from a text prompt. However, current diffusion models do not present a complete solution, as they often require a meticulously crafted prompt, which can be difficult to generate.

In addition, users should have the freedom to make changes to any design that the model generates. Hence, the model must also support a back-and-forth conversation with the user and adapt its responses accordingly. Vision Language Models (VLMs) combine Computer Vision principles with Natural Language Processing enabling them to generate content that has both image and text modalities. For example, Flamingo \cite{alayrac2022flamingo}, a popular VLM, can be trained to do in-context few-shot learning in open-ended tasks, such as visual-question answering. The input to Flamingo is a sequence of arbitrarily interleaved text and images, while its output is only text. It cannot generate any image data. Alternatively, Microsoft's Visual ChatGPT \cite{wu2023visual} has shown that it can generate images, support a conversation with the user, whilst also making changes to its responses based on user requests. It leverages Chain-of-Thought prompting \cite{wei2023chainofthought} in combination with incorporating different Visual Foundation Models (VFMs) with ChatGPT to help process images. However, it needs significant prompt engineering and is limited by the performance of ChatGPT and the various VFMs, all of which affect its performance on any device.

Hence there is a need for multi-modal LLMs. Surveying the current landscape of popular multi-modal LLMs, Meta's ImageBind \cite{girdhar2023imagebind} can encode information across six different modalities including text, audio, and video. It has been shown to generate contextual embeddings across modalities and can also match data across modalities through a similarity score. 
On the other hand, Composable Diffusion (CoDi) \cite{tang2023any} is designed to generate a combination of different modularities given an input of a combination of multiple modularities, e.g., text, image, video, and audio. Another prominent model to consider is OpenAI's GPT-4 \cite{openai2023gpt4} which can process both image and text data. A recent study in evaluating GPT-4V's design generation capability \cite{picard2023concept}, has shown that the model displayed great potential in conceptualizing design but it lacks precision with its responses in later stages of design. Another key finding is that the model often performs better when additional context is provided, and is dependent on prompt engineering for optimal results.

While the current cloud-based models are imperfect tools for engineering design tasks, we hypothesize that these models will improve significantly in the near future, as more data and tools become available. 
No matter the performance, current models can either be deployed on cloud servers and made available to the public via the internet or offline if large computing and memory infrastructures are in place. Neither of which is feasible in rural and remote areas.
This position paper considers how these large multi-modal LLMs can be deployed on edge devices, which will be key when the models become effective design assistants.
\section{Design for the Developing World}

Offline ML models trained on local data have the potential to break through constraints that keep many developing communities from the benefits of global technological advancements. A paper outlining opportunities for mechanical design advancements in the developing world \cite{mattson2016developing} describes how these communities have fundamentally different constraints than those for which many technologies are developed, and that current optimization techniques are catered towards the developed communities they were designed by. One specific example described is sewage systems, which have been present across the world for many years but have not reached many developing communities \cite{mattson2016developing}. These observations lead to the conclusion that there is great potential for the optimization of unique design problems in unique communities. 
\subsection{Predicting Potential Impacts}
Before diving into AI-generated designs, we must also consider the implications of designing for these communities. An article detailing principles for design in the developing world emphasizes co-designing with people from the specific community, testing the product in the actual setting, and adapting technology and product management techniques to the specific context \cite{mattson2014nine}. These principles prioritize participation of the community and customization for the situation. Analysis of failure reports identified the most common pitfall in designing for the developing world as "Lacking the contextual knowledge needed for significant impact" \cite{wood2016design}. Again, this highlights gaining understanding of a community's needs before embarking on design efforts. Collecting this information usually requires direct human interaction, such as interviews and observational studies, but frameworks have proposed combining direct data with indirect data, such as sensors and satellites. These frameworks used deep learning models with a combination of direct and indirect data to predict the social impact of the use of a water hand pump in Uganda \cite{stringham2020combining}. Another study predicted the social impacts of deploying improved cook stoves in Uganda using ML models \cite{mabey2023simulating}. By predicting these social impacts, models can make better-informed decisions about an ideal design. 

When considering design for these communities, we must consider the importance of data collection and customization, as well as the potential contributions that AI can bring to the analysis of this data. More specifically, we can consider how offline AI models can promote participation of a community in their own design decisions. 
\subsection{Product Design and Repair}
AI models have already shown their potential contributions to smart farming systems, which use advanced technologies to facilitate labor reduction and convenience. Using sensor and controller data collected from smart farms, an error detection system using ontology and Recurrent Neural Networks was developed and implemented \cite{choe2023artificial}. 

While these designs have been targeted towards more technologically advanced communities, the process of reverse engineering, where designs are first made for poorer countries and then adapted to wealthier countries, is becoming more globally relevant. In a reverse engineering study, designers were able to meet the specific needs of wheelchair users in the developing world, such as going off road, which were not obvious in the developed world design space \cite{judge2015developing}.

A design activity by Penn State researchers identified low-cost greenhouse designs, with the selection of materials optimized for the requirements of a given location \cite{greenhouse}. Both the development and deployment of these designs was facilitated by outside sources.
Offline AI models provide the opportunity to improve the accessibility of design generation, allowing communities to customize their designs without external support. 

\begin{figure}[hbt!]
    \centering 
    \includegraphics[width=\columnwidth]{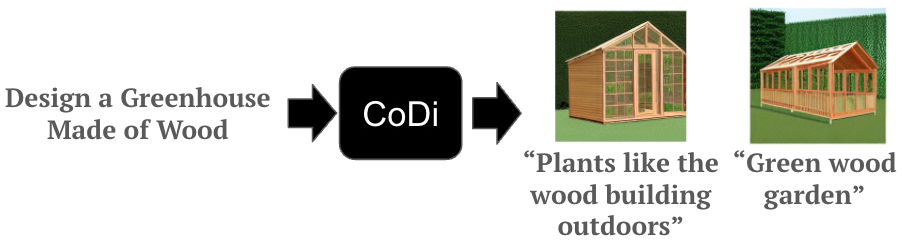}
    \caption{Images and captions generated by CoDi given a wooden greenhouse design request.}
    \label{fig:CoDi_Greenhouse}
\end{figure}

An example of this workflow is demonstrated in Figure \ref{fig:CoDi_Greenhouse}, which shows a user prompting CoDi to "Design a Greenhouse Made of Wood". In an ideal application, the user would be provided with both photos and instructions for design. This example shows some helpful images, but less helpful text outputs. Based on the resources a community has access to, they could utilize AI models to customize their designs. 

\begin{figure}[hbt!]
    \centering 
    \includegraphics[width=\columnwidth]{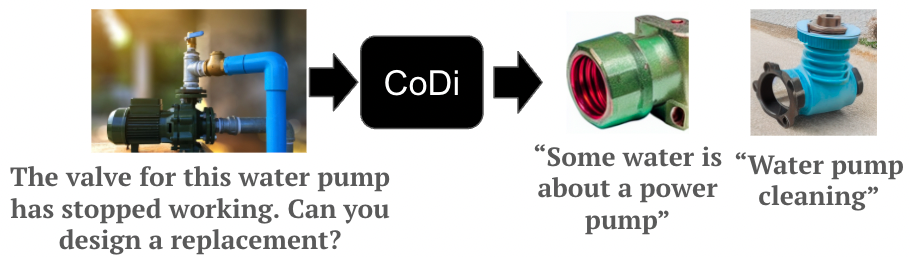}
    \caption{Images and captions generated by CoDi given a design replacement intent.}
    \label{fig:CoDi_Rec_Example_DesignMod}
\end{figure}

Another potential application involves providing those with limited resources the opportunity to repair broken systems quickly and without outside support. 
More specifically, we can return to a sewage system and imagine a water pump stops functioning in a remote area. Instead of waiting for a repairman or a replacement part to arrive, a user could input text with requirements and images of broken components. Light-weight, more accessible models could parse this multi-modal input to propose repair solutions and specific design suggestions in text and image format. An example of this workflow is shown in Figure \ref{fig:CoDi_Rec_Example_DesignMod}, where we inputted an image of a water pump assembly and asked for a replacement valve design. CoDi then generated example images of replacement parts and captions. The current state of this model does not provide particularly helpful design outputs but demonstrates a potentially effective interaction with a user. If trained on specific relevant data, this method would be much more efficient than waiting for outside help or replacement parts, and much more customizable than the design solutions that currently exist for more developed communities. 

Most mechanical design efforts are focused on more resource-heavy communities. Offline models provide an opportunity to customize a generated design to a given set of constraints.
\section{Potential Solutions}
In this section, we discuss some of the ways one can address the challenges faced by resource-constrained areas in accessing large AI models like Language Models (LLMs). 

The most common solution is fine-tuning for Cloud-based solutions as some resource-constrained areas may have internet access. In such areas, users can access cloud services with minimal local computing resources. Efforts to optimize data transfer and enable low-bandwidth operation are key in this approach. With in-context learning, the models can potentially be adapted to local needs. However, in this paper, we do not focus on these cloud-only solutions. Instead, our focus is on 
offline AI solutions.

Academics and industry are contributing to open-source projects and community-driven initiatives that aim to make AI more accessible. This includes sharing resources, providing training, and supporting local developers in creating solutions tailored to their specific needs. However, these open source models could still be large and need significant compute resources to train. 
Below, we discuss a few key solutions to make them more accessible and suitable for deployment on the edge.

\subsection{Model Compression and Optimization}
 Deploying AI models on edge devices is far from straightforward. Due to memory and computing constraints, not all models can be run on the edge. There is a need for more efficient models that require less computational power. This is even more pronounced when dealing with LLMs. Their higher memory and processing requirements aren't practical for computing at the edge, thus making them less accessible. One of the ways to achieve this is to make these models smaller for inference using methods like:
\begin{itemize}
    \item
    \textbf{Model Pruning:} The process of removing non-critical and redundant components of a model without a significant loss in performance. With respect to LLMs, this can mean removing weights with smaller gradients or magnitudes and parameter reduction among others. Novel pruning methods like Wanda \cite{sun2023simple} and LLM-Pruner \cite{ma2023llmpruner} present optimal solutions for making LLMs smaller.

    \item
    \textbf{Quantization:} Representing model parameters such as weights in a lower precision, i.e. using fewer bits to store the value \cite{gholami2021survey}. This results in a smaller model size, faster inference, and a reduced memory footprint. LLM Quantization can be achieved either in the post-training phase  \cite{dettmers2022llmint8} or during the pre-training or fine-tuning phase \cite{liu2023llmqat}.
    
    \item
    \textbf{Knowledge Distillation:} Transferring the knowledge of a large teacher model to a smaller learner model to replicate the original model's output distribution difference. Knowledge Distillation has been widely used to reduce LLMs like BERT into smaller distilled versions DistilBERT \cite{sanh2020distilbert}. More recently, approaches like MiniLLM \cite{gu2023knowledge} and \cite{hsieh2023distilling} further optimize the distillation process to improve the student model's performance and inference speed.
\end{itemize}

\subsection{Edge Computing}
Edge computing involves processing data closer to the location where it is needed, rather than relying on a central data-processing warehouse. This can significantly reduce the need for continuous, high-speed internet connectivity. One solution is to deploy AI models on edge devices, like smartphones and local servers, which can operate with intermittent connectivity \cite{singh2023edge, radu2020edgeai}.

\subsection{Evaluation Metrics}
Depending on the design objective and types of constraints, metrics can be defined to evaluate the performance of a model. 
Selection of these metrics is crucial, but can also vary greatly depending on the representation
scheme, the cost of metric evaluation, and the use case.
Examples of common metrics within the 3D design space can include statistical similarity to highly performing designs, diversity in design candidates and design performance evaluations \cite{regenwetter2023beyond}.
When applied to offline models, we will need to consider metrics that have low-resource evaluation capabilities and prioritize low cost and performance in designs. 
\\

Other solutions for the lack of computational resources also exist. For example, the development of custom hardware, like specialized AI chips, can offer a more cost-effective and power-efficient way of running AI models. These chips can be integrated into everyday devices, reducing the need for powerful standalone computers. There's also an emerging trend of using multiple devices to share computational tasks. This distributed approach can help overcome individual device limitations. Finally, there is also a need for establishing data centers closer to resource-constrained areas to reduce latency and improve access. This approach, however, requires significant investment and infrastructure development.
\section{Discussion: Edge AI and Tiny ML}
As discussed in the Review section, existing approaches like LLMs, Diffusion models and VLMs have significant computing requirements owing to their large parameter space which make them imperfect candidates for deployment on edge devices. On top of the optimization and compression techniques mentioned in the Potential Solutions section, more work needs to be done both on the hardware and software side to increase their accessibility \cite{lin2023pushing}.

Recent research in Edge AI \cite{Murshed_2021, lee2018edgeai} has shown that optimized model architectures and new embedded system technologies improved system response times as well as user security. Similarly, TinyML is a field of Machine Learning that deals with deploying lighter models in resource-constrained devices that have limited computing infrastructure. Certain key aspects of TinyML include reducing latency by leveraging edge computing, optimizing models for computing at the edge and developing new tools and equipment that can facilitate running ML models on edge devices. These features enable TinyML models to provide increased user privacy due to data being processed locally, reduced power consumption and internet usage \cite{dutta2021tinyml, ray2022review}.

The findings from these both of these fields are crucial in realizing the goal of democratizing Generative AI for computing at the edge.
\section{Conclusion}

With this position paper, we want to continue the discussion on making Generative AI models more accessible and increasing their reach to rural and remote areas. Researchers developing solutions for data-driven design challenges must look to create models that can also work in resource-constrained environments, not just on the cloud. The future of AI in design at the edge is not just about technological innovation but also about equitable access and sustainability. It is about bringing the power of AI-driven design to every corner of the globe, ensuring that even the most remote communities can benefit from the advancements in AI for engineering and design solutions.

\bibliography{aaai24}

\end{document}